\documentclass[lettersize,journal]{IEEEtran}
\usepackage{amsmath,amsfonts}
\usepackage{algorithmic}
\usepackage{algorithm}
\usepackage{array}
\usepackage[caption=false,font=normalsize,labelfont=sf,textfont=sf]{subfig}
\usepackage{textcomp}
\usepackage[]{changes}
\usepackage{stfloats}
\usepackage{url}
\usepackage{verbatim}
\usepackage{graphicx}
\usepackage{cite}
\usepackage{bm, xcolor}
\usepackage{tikz}
\usepackage{tabularx}
\usepackage{booktabs}
\usetikzlibrary{calc}
\usetikzlibrary{arrows.meta}
\usetikzlibrary{positioning, backgrounds, fit, shapes}

\newcommand{\ABLSDG}{DSVDD-KL}
\newcommand{\ABLSDL}{DLSVDD}
\newcommand{\ABLSDLG}{DLSVDD-KL}
\newcommand{\ABLAEG}{IAE-KL}
\newcommand{\ABLAEL}{IAE-LSTM}
\newcommand{\MODEL}{IAE-LSTM-KL}
\newcommand{\MODELEXT}{Improved AutoEncoder with LSTM module and KL divergence}
\newcommand{\cifar}{{CIFAR10}}%~\cite{krizhevskycifar10}}}

\begin{document}

\title{{Improved AutoEncoder with LSTM Module and KL Divergence for Anomaly Detection}}

\author{Wei Huang, Bingyang Zhang, Kaituo Zhang, Hua Gao and Rongchun Wan
    % <-this % stops a space
    \thanks{This work was supported by the National Key R\&D Program of China (2022YFE0198900) and the National Natural Science Foundation of China (61771430).}
    \thanks{Wei Huang, Bingyang Zhang, Kaituo Zhang and Hua Gao are with the College of Computer Science, Zhejiang University of Technology, Hangzhou, 310023, China.}
    \thanks{Rongchun Wan is with Zhejiang HOUDAR Intelligent Technology Co., Ltd., Hangzhou, 310023, China.}
    \thanks{Hua Gao is the corresponding author (e-mail: ghua@zjut.edu.cn).}}
\markboth{IEEE Transactions on Instrumentation and Measurement,~Vol.~14, No.~8, August~2021}%
{Shell \MakeLowercase{\textit{et al.}}: A Sample Article Using IEEEtran.cls for IEEE Journals}

\maketitle
\begin{abstract}
    The task of anomaly detection is to separate anomalous data from normal data in the dataset. Models such as deep convolutional autoencoder (CAE) and deep supporting vector data description (SVDD) have been universally employed and have demonstrated significant success in detecting anomalies. However, the over-reconstruction ability of CAE network for anomalous data can easily lead to high false negative rate in detecting anomalous data. On the other hand, the deep SVDD model has the drawback of feature collapse, which leads to a decrease of detection accuracy for anomalies. To address these problems, we propose the Improved AutoEncoder with LSTM module and Kullback-Leibler divergence (IAE-LSTM-KL) model in this paper. An LSTM network is added after the encoder to memorize feature representations of normal data. In the meanwhile, the phenomenon of feature collapse can also be mitigated by penalizing the featured input to SVDD module via KL divergence. The efficacy of the IAE-LSTM-KL model is validated through experiments on both synthetic and real-world datasets. Experimental results show that IAE-LSTM-KL model yields higher detection accuracy for anomalies. In addition, it is also found that the IAE-LSTM-KL model demonstrates enhanced robustness to contaminated outliers in the dataset.
\end{abstract}

\begin{IEEEkeywords}
    LSTM, Deep SVDD, autoencoder, hypersphere collapse, anomaly detection.
\end{IEEEkeywords}
\section{Introduction}
\label{sec:introduction}
\IEEEPARstart{I}{n} recent years, the topic of anomaly detection has emerged as a popular research topic in both science and application fields.
The applications of anomaly detection span across various domains, which include intrusion detection systems of networks \cite{shone2018deep}, fraud detection systems \cite{huang2018codetect} and so on.
To address the task of anomaly detection, unsupervised learning methods have been widely employed in recent years.
In contrast to supervised or semi-supervised learning, unsupervised learning has the advantage of better suitability to situations where labeled or categorized information is scarce \cite{zhang2023destseg, zhang2017split,golan2018deep}. Therefore, we mainly focus on unsupervised anomaly detection in this study.

To cope with the task of anomaly detection, the Convolutional AutoEncoder (CAE) \cite{masci2011stacked} model has gained significant attention for its ability to learn compressed, meaningful representations of high-dimensional data.
The common assumption of CAE is that the reconstruction error tends to be lower for normal inputs, and is anticipated to increase for abnormal inputs.
However, the CAE model has the drawback that this model can sometimes reconstruct anomalous inputs very well, thus resulting in lower reconstruction error for anomalous inputs.
    {To further improve discriminate ability of CAE model for normal and abnormal inputs, researchers then proposed a few other variants of CAE such as Variational AutoEncoder (VAE) \cite{kingma2013auto} and Denoising AutoEncoder (DAE) \cite{vincent2010stacked}  to enhance the power of autoencoder network in anomaly detection. The VAE model offers performance improvement over CAE by generating smooth, continuous samples with structured latent space, providing probabilistic modeling for uncertainty estimation and facilitating meaning interpolation in the latent space.  The purpose of DAE \cite{vincent2010stacked}  model is to add noise to input data to enhance the performance of CAE by strengthening the robustness of learned representations to partial corruption of input data. In addition, in \cite{NIPS2017_7a98af17}, the authors proposed the Vector Quantised-Variational AutoEncoder (VQ-VAE), in which the encoder outputs representations from the distribution over discrete variables. Experimental results in \cite{NIPS2017_7a98af17} demonstrate that the discrete latent space learnt by VQ-VAEs can capture important features of input data in completely unsupervised manner. In addition, to improve the discriminative capacity of CAE, \cite{gong2019memorizing} incorporated a memory module to confine the latent space of an autoencoder in the context of video anomaly detection.}

%To further improve discriminative ability of CAE model for normal and abnormal inputs, researchers then proposed a few other variants of CAE such as Variational Autoencoder (VAE) \cite{kingma2013auto} to enhance the power of autoencoder network in anomaly detection.
%The VAE model offers performance improvement over the CAE by generating smooth, continuous samples with a structured latent space, providing probabilistic modeling for uncertainty estimation, and facilitating meaningful interpolation in the latent space.
%{Inspired by VAE, in \cite{NIPS2017_7a98af17}, the codebook structure was inserted into the latent space of CAE, replacing the continuity of discrete vector encoding in the latent space. The model in \cite{NIPS2017_7a98af17} was called Vector Quantised-Variational AutoEncoder (VQ-VAE), based on the theory that discrete vector encoding improves the performance of CAE compared to continuous encoding.
%    The purpose of the DAE \cite{vincent2010stacked} is to add noise to images to improve the outcomes of CAE by making the learned representations robust to partial corruption of the input patterns.}
%In addition, to improve the discriminative capacity of CAE, \cite{gong2019memorizing} incorporated a memory module to confine the latent space of an autoencoder in the context of video anomaly detection.

% In \cite{rippel2021gaussian}, the authors proposed a novel method, called Gaussian AD method, to detect anomalies by modelling a multivariate Gaussian (MVG) distribution for the normal data in deep feature representations.
Support Vector Data Description (SVDD) \cite{tax2004support} represents another kind of unsupervised learning designed for one-class classification.
The primary objective of SVDD is to characterize the inherent structure of data samples by enveloping the majority of data points within a hypersphere in the high dimensional space.
However, SVDD cannot work well in high-volume and high-dimensional dataset such as image or video dataset.This is mainly because SVDD is a kind of shallow method which cannot sufficiently capture intrinsic deep features in high-dimensional space.
To address this problem, Deep Support Vector Data Description (Deep SVDD) \cite{ruff2018deep}, an unsupervised deep-learning approach for one-class classification, has been proposed recently.
This approach is inspired by the SVDD model \cite{tax2004support} and reduces the data dimension into an appropriate size which is realized by an encoder module, mapping the original high-dimensional data into feature vectors lying in the closed hypersphere space.
However, \cite{ruff2018deep} also pointed out that Deep SVDD has the problem of hypersphere collapse, in which all data points are mapped to a single point in feature space, losing the ability to distinguish anomalous data from normal data.
This problem can be partially overcome by removing bias terms of all hidden layers since the network with the bias term in any hidden layer can be trained into a constant function mapping all data into the hypersphere center, leading to hypersphere collapse \cite{ruff2018deep}.
{Another idea to mitigate the phenomenon of feature collapse in Deep SVDD is to combine Deep SVDD with the AutoEncoder. In \cite{cheng2021improved}, the authors pointed out that by learning embedded features from input data, the AutoEncoder model possesses the mechanism of local structure preservation. This mechanism enables the AutoEncoder model to partially overcome the existing drawback of Deep SVDD model. Thus, the combination of AutoEncoder and Deep SVDD can effectively alleviate the phenomenon of feature collapse in the feature space.}

{To further address the problem of over-reconstruction ability for anomalous data in CAE model and phenomenon of feature collapse in Deep SVDD model, we propose the Improved AutoEncoder with LSTM module and KL divergence (called IAE-LSTM-KL for short) approach in this paper. Since this model is the composition of AutoEncoder, Deep SVDD, LSTM and KL divergence, it is essential to make some explanations why we adopt the combination strategy. Firstly, as we have stated in the paragraph above, the combination of AutoEncoder and Deep SVDD is to alleviate the phenomenon of feature collapse in the feature space. Secondly, the use of KL divergence serves as a penalty to promote the latent vectors towards the Gaussian distribution. By using KL divergence, the feature vectors fed into the SVDD module is shifted from the center of hypersphere towards outer periphery of hypersphere in the hidden space, thus reducing the possibility of feature collapse in Deep SVDD module. Finally, considering that the LSTM module excels in capturing temporal features of time-series data, we add an LSTM layer between the encoder and the decoder to store feature representations learned in the training stage. The LSTM layer has the role of memorizing feature representations from normal samples. The input gate and the output gate in LSTM module can filter out the impact of abnormal information by discerning the flow of information, ensuring the purity of normal data.}

The main contributions of this study are summarized as follows:
\begin{itemize}
    \item We propose a novel anomaly-detection model, called Improved AutoEncoder with LSTM module and KL divergence (\MODEL{}) in this paper, to simultaneously address the problem of over-reconstruction for anomalous data in CAE model and the problem of feature collapse in Deep SVDD model. This model is the combination of the autoencoder framework, the SVDD module, the LSTM module and the KL divergence penalty. Different from the one-class Deep SVDD and CAE models, the \MODEL{} model can preserve essential normal features while concurrently sieving out anomalous data, thus diminishing the total loss relevant to normal cases.
    \item Extensive experiments have been carried out on multiple datasets, including both synthetic and real-world datasets. Results show that the proposed \MODEL{} model exhibits superior accuracy and reduced error rates as compared to other state-of-art anomaly-detection models.
    \item  Experimental results also show that the \MODEL{} model demonstrates enhanced robustness to contaminated noises in the dataset. Even though the training set is contaminated with some outliers, the \MODEL{} model can still retain high capacity to detect anomalies, thereby endorsing the consistent enhanced robustness to noises in the dataset.
\end{itemize}

The rest of this paper is organized as follows.
In Section II, we describe some preliminary work before formally introducing our proposed IAE-LSTM-KL model.
In Section III, we describe the proposed IAE-LSTM-KL model in detail.
In Section IV, the performances of the IAE-LSTM-KL model and several competing models in anomaly detection are compared through experiments on some datasets. Extensive discussions for experimental results are also presented in this section.
Finally, some conclusions are drawn in Section V.
All code and hyperparameters may be found at https://github.com/crazyn2/IAE-LSTM-KL\_codes

{\textbf{Notations}: Bold-faced uppercase and lowercase letters denote matrices and vectors respectively. $ \odot $ denotes the operation of element-wise multiplication between two vectors with the same dimension. $\left\|  \bullet  \right\| $ denotes the empirically measured $l_2$-norm of argument vector. ${\left\|  \bullet  \right\|_F}$ denotes the Frobenius norm of the argument matrix. $\sigma ( \bullet )$ denotes the sigmoid function. $\tanh ( \bullet )$ denotes the hyperbolic tangent function. $KL( \bullet \parallel  \bullet )$ denotes the Kullback-Leibler divergence of two probability distributions. Finally, ${\mathcal{N}_{{\mathbf{0,I}}}}$ denotes the standard Gaussian distribution.}

\section{Related Work}
\label{sec:related}
Before presenting our proposed model, in this section, we describe several prior works that are relevant to our work.
\subsection{Convolutional AE (CAE)}
The convolutional autoencoder (CAE) is a kind of unsupervised reconstruction-based neural network, which consists of an encoder $f_{e}(\cdot ; \theta_{e})$ and a decoder $f_{d}(\cdot ; \theta_{d})$.
The encoder compresses input data into the lower-dimensional latent feature, while the decoder attempts to reconstruct the original data from the compressed representation.
With the assumption that the CAE model is usually trained on normal data, the training objective is that abnormal inputs result in a larger reconstruction error than normal inputs in the testing phase.
Specifically speaking, the dependency between the input data $\mathbf{x}$ and the reconstructed signal $\mathbf{\hat{x}}$ can be expressed in the following equation:
\begin{equation}
    \mathbf{\hat{x}} = f_{d}(f_{e}(\mathbf{x}; \theta_{e}); \theta_{d}).
    \label{eq:autoencoder}
\end{equation}
In addition, the target loss function of the CAE model is to minimize
\begin{equation}
    L = \min_{\theta_e, \theta_d}\|\mathbf{x} - \mathbf{\hat{x}}\|,
    \label{eq:autoencoder_loss}
\end{equation}

\subsection{Deep SVDD}
The Deep SVDD \cite{ruff2018deep} model is a kind of one-class anomaly detection method which utilizes the encoder part of autoencoder.
The training objective of Deep SVDD is to map normal samples into a constructed hypersphere with a center in the high-dimensional space.
In the testing phase, normal samples fall within the hypersphere, while anomalous samples fall outside the hypersphere.

Define the input space $\mathcal{X} \subseteq R^d$ and hidden feature space $\mathcal{F}$. Given the input data $\mathbf{x} \in \mathcal{X}$, the expression $f_e(\cdot ; \theta_{e}) :\mathbf{x} \to \mathbf{z}$ denote an encoder network with $L$ hidden layers such as convolutional neural layer for image dataset, and a set of weights $\mathcal{W} = \{\mathbf{W}^1,...,\mathbf{W}^L\}$.
Here, $\mathbf{z} \in \mathbf{\mathcal{F}}$ denotes the feature representations of $\mathbf{x} \in \mathbf{\mathcal{X}}$ in the hidden space.
The aim of Deep SVDD is to jointly learn the network parameters $\mathbf{\mathcal{W}}$ together with minimizing the volume of a data-enclosing hypersphere in the feature space $\mathbf{\mathcal{F}}$.
The hypersphere is characterized by radius $R > 0$ and center $\mathbf{c} \in \mathcal{F}$.
Given the training set $D_n=\{\mathbf{x}_1,...,\mathbf{x}_n\}$ on $\mathcal{X}$, \cite{ruff2018deep} defines two different kinds of objective functions. The first is the soft-boundary Deep SVDD:
\begin{equation}
    \begin{aligned}
        L_{soft} = & \min_{\theta_e, R, \mathcal{W} } R^2+\frac{1}{\upsilon n} \sum_{i=1}^{n} max(0, \|\mathbf{z_i} - \mathbf{c}\|^2-R^2) \\
                   & +\frac{\lambda}{2} \sum_{l=1}^{L} \|\mathbf{W}^l\|_{F}^2,                                                            \\
    \end{aligned}
    \label{eqn: soft}
\end{equation}
and the second is the hard-boundary Deep SVDD:
\begin{equation}
    \begin{aligned}
         & L_{hard}= \min_{\theta_e, \mathcal{W} } \frac{1}{n}\sum_{i=1}^{n}\|\mathbf{z_i} - \mathbf{c}\|^2+\frac{\lambda}{2} \sum_{l=1}^{L} \|\mathbf{W}^l\|_{F}^2. \\
    \end{aligned}
    \label{eqn: hard}
\end{equation}
The last terms of both \eqref{eqn: soft} and \eqref{eqn: hard} are the weight decay regularizers to alleviate overfitting for the input data with the hyperparameter $\lambda > 0$. In addition, the hyperparameter  $\upsilon$ in \eqref{eqn: soft} controls the degree of some points being mapped out of the hypersphere.
The hard-boundary objective function \eqref{eqn: hard} can be regarded as the simplified version of soft-boundary objection function when the most training data is normal. In our study, we {will} apply both soft-boundary and hard-boundary versions in our proposed \MODEL{} model.
% \begin{figure}
%     \begin{tikzpicture}
%         \begin{axis}[axis equal]
%             \addplot[
%                 scatter,
%                 only marks,
%                 samples=50,
%                 point meta=100-sqrt(x^2+y^2),
%                 coordinate style/.condition=
%                     {sqrt(x^2+y^2) < 0.5}{scale=2},
%                 coordinate style/.condition={\coordindex == 25}{
%                         /pgfplots/scatter/@post marker code/.add code={
%                                 \node [pin=45:anomaly] {};
%                             }{},
%                     },
%             ] (rand,rand);
%             \draw (0,0) circle (0.5);
%         \end{axis}
%     \end{tikzpicture}
%     \caption{Graphical representation of low-dimensional latent vector mapping in the context of the Deep SVDD or SVDD algorithm.} \label{fig:svdd}
% \end{figure}
% \begin{figure*}
%     \includegraphics[width=0.9\linewidth]{figures/deepsd}
%     \caption{Graphical representation of low-dimensional latent vector mapping in the context of the Deep SVDD algorithm.} \label{fig:svdd}
% \end{figure*}

\subsection{Improved AutoEncoder for Anomaly Detection (IAEAD)}
As we have stated above, both CAE and Deep SVDD models have the capability of detecting anomalies.
However, in the case that the training data is contaminated by some anomalous data, the CAE model can well reconstruct both normal and anomalous data, which leads to high false negatives in anomaly detection.
On the other hand, the Deep SVDD model has the drawback of easily suffering from feature collapse for anomaly detection, which mixes hidden features of both normal and abnormal samples together and results in indistinguishability of anomalous samples from normal samples.

To overcome the drawbacks of CAE and Deep SVDD models, \cite{cheng2021improved} proposed an improved autoencoder model (called IAEAD) that combines CAE with Deep SVDD.
This improved model can efficiently implement the task of anomaly detection while learning representative features to preserve the local structure of input data.
Specifically speaking, the IAEAD model is optimized by minimizing the following objective function:
\begin{equation}
    \begin{aligned}
         & L = \min_{\mathcal{W}} \alpha L_{rec} + L_{soft}. \\
        %  & L = \min_{\mathcal{W}} \lambda L_{rec} + L_{oc}\ \text{or}\ L_{soft} \\
    \end{aligned}
    \label{eq: aesd}
\end{equation}
As we can observe from \eqref{eq: aesd}, the objective function of IAEAD model is actually the weighted sum of $L_{rec}$ and $L_{soft}$ controlled by the hyperparameter $\alpha$.
Finally, we have to mention that the term $L_{soft}$ can also be replaced by $L_{hard}$, forming the IAEAD model with hard-boundary Deep SVDD model.

\section{Proposed Method}
\label{sec:method}
\subsection{Framework of \MODEL{} model}
\begin{figure}[!t]
    \centering
    \includegraphics[width=0.75\linewidth]{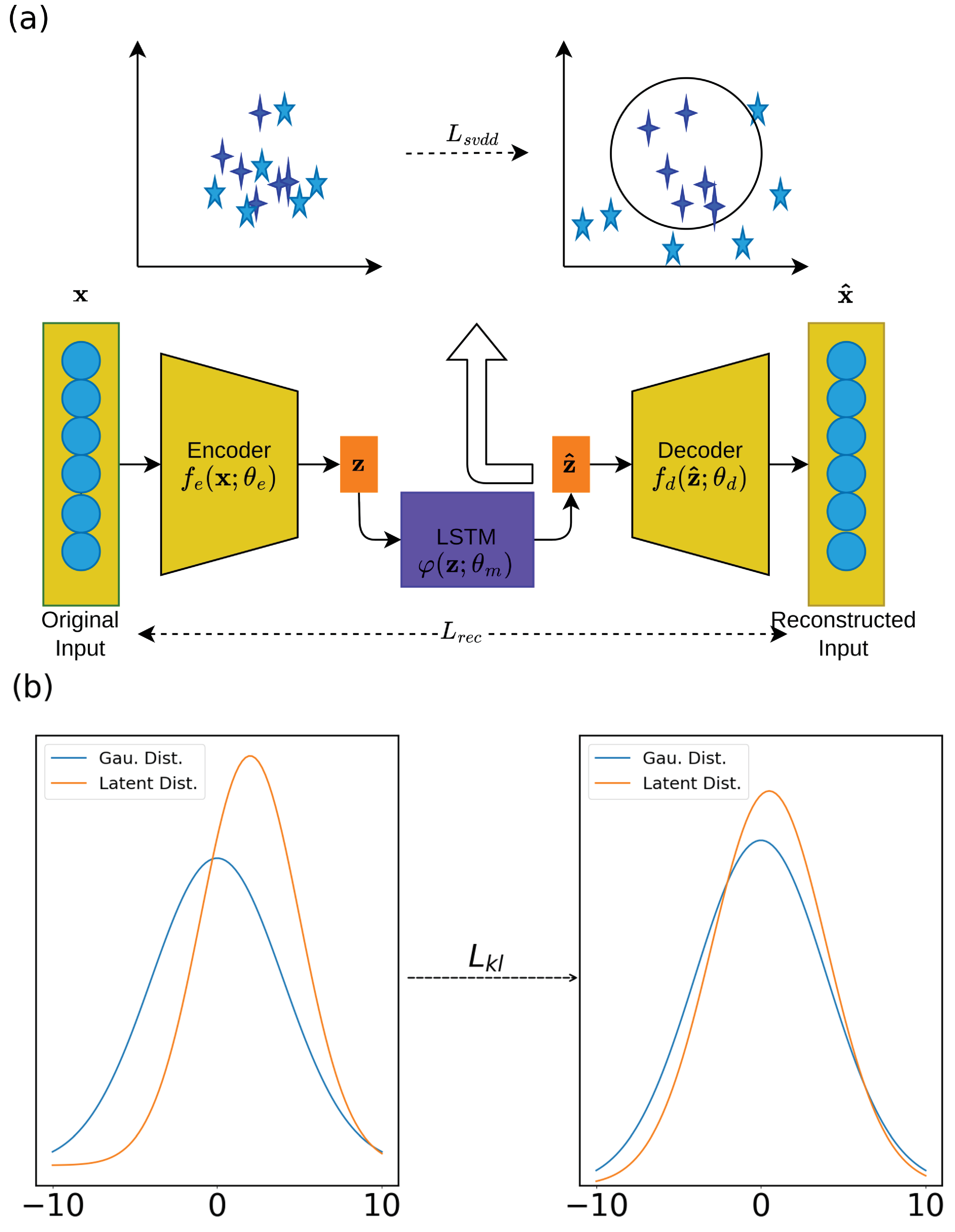}
    \caption{(a). Schematic representation of the whole IAE-LSTM-KL model. (b). The role of KL divergence in forcing latent feature vectors following standard Gaussian distribution.} \label{fig:arch}
\end{figure}

The framework of \MODEL{} model is illustrated in Fig.~\ref{fig:arch}(a).
This model consists of three modules, i.e. the CAE module, the SVDD module and the LSTM module.
The encoder in CAE module initially processes the input data to generate the latent feature vector in the feature space.
Subsequently, the LSTM module utilizes this latent vector to capture the feature relationships between the input data and the stored normal data representatives.
Then, the decoder in CAE module reconstructs the latent features generated from the LSTM module.
Simultaneously, the role of SVDD module is to create a minimal hypersphere enclosing most latent features generated from the LSTM module.

As we have mentioned in the section of Introduction, the Deep SVDD model \cite{ruff2018deep} has the drawback of feature collapse in the training process.
Hypersphere collapse hinders the model to distinguish abnormal data samples from normal samples, leading to difficulties in detecting anomalies.
To further mitigate the phenomenon of feature collapse in Deep SVDD module, we use Kullback-Leibler (KL) divergence to force the latent features after the LSTM module following Gaussian distribution (shown in Fig.~\ref{fig:arch}(b)), which can provide more flexibility for latent features after the LSTM module to cover broader range of latent representation space.
By using KL divergence to regulate the distribution of latent features, the proposed model can well preserve the diversity of latent features fed into the Deep SVDD module.
Consequently, the robustness of the proposed \MODEL{} model in detecting anomalies can be enhanced.

In the training phase, the \MODEL{} model is assumed to be exclusively trained using normal data.
The LSTM module continuously updates the contents stored in it, and memorizes the prototype representations of normal data.
In the testing phase, even when an abnormal input is sent into the \MODEL{} model, the reconstructed data still resembles normal samples resulting in high reconstruction error since the reconstructed data is decoded from normal patterns in LSTM.
As a result, the reconstruction error is still nontrivial if the input data is abnormal.
In the meanwhile, the reconstruction error still remains to be low for normal input.
Therefore, the LSTM module in the \MODEL{} model plays an important role in distinguishing anomalous data from normal inputs.

In the following part, we will firstly analyze the working mechanism of LSTM module. Then, we {will} give the overview of the whole \MODEL{} in detail.

\subsection{Introduction for LSTM module}
LSTM \cite{hochreiter1997long} is a type of memory that extracts similar information from the input data and stores them in it.
Usually, a LSTM network consists of multiple cell units and the structure of each cell unit is demonstrated in Fig.~\ref{fig:lstm}.
In the LSTM structure, {$\mathbf{c}_{i,t}$} represents the cell state, which is used as the memory unit storing information from the previous cell unit and receiving information from current cell unit.
In each cell unit, there are three gates, namely the input gate {$s$}, the output gate $o$ and the forget gate $f$ respectively.
The input gate controls the flow of information into the cell state {$\mathbf{c}_{i,t}$} which contains the stored representatives of normal data.
The role of input gate is to determine what to remember from the current input.
The output gate decides what to output as the new hidden state using the sigmoid function.
Finally, the forget gate regulates what to forget from the previous cell state.

\begin{figure}
    \centering
    %    \subcaptionbox{}[.5\textwidth]
    {
        \tikzset{global scale/.style={
                scale=#1,
                every node/.append style={scale=#1}
            }
    }
    \begin{tikzpicture}[
        % 定义样式
        % scale=0.8,
        global scale =0.8,
        % \tikzstyle{every node}=[font=\small,scale=0.8],
        myfont/.style={
                font=\footnotesize\sffamily
            },
        frame/.style={% 外部边框(圆角矩形)的样式
                rectangle,
                rounded corners=5mm,
                draw,gray,
                fill=lime!45,
                very thick,
            },
        add/.style={% 加操作符的样式
                circle,
                draw, thick,
                fill=yellow!60,
                inner sep=.5pt,
                minimum height =3mm,
            },
        prod/.style={% 乘操作符的样式
                circle,
                draw, thick,
                fill=lightgray!40!blue!30,
                inner sep=.5pt,
                minimum height =3mm,
            },
        function/.style={% 信息处理函数(tanh)的样式
                ellipse,
                draw, thick,
                fill=cyan!30,
                inner sep=1pt
            },
        cell/.style={%C_t和C_{t-1}的样式
                circle,draw,fill=magenta!30,
                line width = .75pt,
                minimum width=8mm,
                inner sep=1pt,
            },
        hidden/.style={% h_t和h_{t-1}的样式
                circle,draw,fill=orange!50,
                line width = .75pt,
                minimum width=8mm,
                inner sep=1pt,
            },
        input/.style={% z_t的样式
                circle,draw,
                fill=teal!30,
                line width = .75pt,
                minimum width=8mm,
                inner sep=1pt,
            },
        actfunc/.style={% 激活函数的样式
                rectangle, draw,
                fill=pink!30,
                minimum width=4.25mm,
                minimum height=3.75mm,
                inner sep=1pt,
                thick,
            },
        ArrowC1/.style={% 线段的圆角样式
        rounded corners=2mm,
        >=Stealth[round],
        very thick,
        },
        ArrowC2/.style={% 弧度略大的圆角样式
        rounded corners=3mm,
        >=Stealth[round],
        very thick,
        },
        ]

        % 图形绘制部分  
        % 绘制外部圆角边框
        \node [frame, minimum height =4cm, minimum width=6cm] at (0,0) {};

        % 绘制激活函数结点
        \node [actfunc] (ibox1) at (-2,-0.8) {$\sigma$};
        \node [actfunc] (ibox2) at (-1.35,-0.8) {$\sigma$};
        \node [actfunc, minimum width=9mm] (ibox3) at (-0.45,-0.8) {$\tanh$};
        \node [actfunc] (ibox4) at (0.5,-0.8) {$\sigma$};

        % 绘制操作符结点
        \node [prod] (mux1) at (-2,1.5) {$\bm\times$};
        \node [add] (add1) at (-0.5,1.5) {$\bm+$};
        \node [prod] (mux2) at (-0.5,0) {$\bm\times$};
        \node [prod] (mux3) at (1.5,0) {$\bm\times$};
        \node [function] (func1) at (1.5,0.75) {$\tanh$};

        % 绘制神经元外部输入节点
        \node[cell, label={[myfont]above:}] (c) at (-4,1.5) {$\mathbf{c}_{i,t-1}$};
        \node[hidden, label={[myfont]above:}] (h) at (-4,-1.5) {$\mathbf{h}_{i,t-1}$};
        \node[input, label={[myfont]right:Input}] (x) at (-2.5,-3) {$\mathbf{z}_{i,t}$};

        % 绘制神经元外部输出节点
        \node[cell, label={[myfont]above:}] (c2) at (4,1.5) {$\mathbf{c}_{i,t}$};
        \node[hidden, label={[myfont]below:}] (h2) at (4,-1.5) {$\mathbf{h}_{i,t}$};
        \node[hidden, label={[myfont]left:Output}] (x2) at (2.5,3) {$\mathbf{\hat{z}}_{i,t}$};

        % 绘制各节点之间的连接线
        % 使用相交和位移
        % 绘制C_{t-1}到C_t的线
        \draw [->, ArrowC1] (c) -- (mux1);
        \draw [->, ArrowC1] (mux1) -- (add1);
        \draw [->, ArrowC1] (add1) -- (c2);
        % \draw [->, ArrowC1] (c) -- (mux1) -- (add1) -- (c2);
        % \draw [->, ArrowC1] (c) -- (mux1) -- (add1) -- (c2);

        % 绘制输入线段
        \draw [ArrowC2] (h) -| (ibox4);
        \draw [ArrowC1] (h) -| (ibox1);
        \draw [ArrowC1] (h) -| (ibox2);
        \draw [ArrowC1] (h) -| (ibox3);
        \draw [ArrowC1] (x) -- (x |- h) -| (ibox1);

        % 内部线段(带箭头)
        \draw [->, ArrowC2] (ibox1) -- (mux1);
        \node [label=left:$\mathbf{f}_{i,t}$,right=1.5pt] () at ($(ibox1)!.5!(mux1)$) {};
        \draw [->, ArrowC2] (ibox2) |- (mux2);
        \node [label=left:$\mathbf{s}_{i,t}$, right=2pt] () at (ibox2 |- mux2) {};
        \draw [->, ArrowC2] (ibox3) -- (mux2);
        \node [label=right:$\mathbf{\tilde{c}}_{i,t}$, left=.5pt] () at ($(ibox3)!.5!(mux2)$) {};
        \draw [->, ArrowC2] (ibox4) |- (mux3);
        \node [label=left:$\mathbf{o}_{i,t}$, right=2pt] () at (ibox4 |- mux3) {};
        \draw [->, ArrowC2] (mux2) -- (add1);
        \draw [->, ArrowC1] (add1) -| (func1);
        \draw [->, ArrowC2] (func1) -- (mux3);

        %输出线段
        \draw [->, ArrowC2] (mux3) |- (h2);
        %标记出现缺口线段的下端点
        \draw (c2 -| x2) ++(0,-0.15) coordinate (i1);
        \draw [-, ArrowC2] (h2 -| x2) -| (i1);
        \draw [->, ArrowC2] (i1)++(0,0.3) -- (x2);
    \end{tikzpicture}}
    \caption{Demonstration of cell unit in the LSTM network.} \label{fig:lstm}
\end{figure}
% Considering the batch of input data $\{\mathbf{x}_1,\mathbf{x}_2,...,\mathbf{x}_n\} \in \mathcal{X}$, where $n$ denotes the total number of input samples,
{Given the batch latent feature of encoder output $\{\mathbf{z}_1,\mathbf{z}_2,...,\mathbf{z}_n\} \in \mathcal{Z}$, where $n$ denotes the total number of samples, we have to decide what new information we are going to store in the cell state and establish the relationship between $\mathbf{z}_{i,t}$ and the stored cell state $\mathbf{c}_{i,t}$.
In this case, the variable $t$ not only represents the number of LSTM units but also denotes the temporal index of $\mathbf{z}_{i}$ within its tensor structure.}
First, the input gate and the output gate respectively control the flow of information from {$\mathbf{z}_{i,t}$} and the flow of information output from {$\mathbf{c_{i,t}}$}, which can be expressed as
\begin{equation}
    \mathbf{s}_{i,t} = \sigma(\mathbf{W}_{s} \cdot [\mathbf{h}_{i,t-1}, \mathbf{z}_{i,t}] + \mathbf{b}_{s})
\end{equation}
and
\begin{equation}
    \mathbf{o}_{i,t} = \sigma(\mathbf{W}_{o} \cdot [\mathbf{h}_{i,t-1}, \mathbf{z}_{i,t}] + \mathbf{b}_{o})
\end{equation}
respectively. {The signal $\mathbf{h}_{i,t}$ represents the output vector from the $t^{th}$ LSTM unit.} Here, through dual control of input and output, we further refine the information flow to ensure the purity of the regular information stream, thus filtering out anomalous data.
In addition, the forget gate $\mathbf{f}_{i,t}$ is determined by both the {$\mathbf{h}_{i,t-1}$ and $\mathbf{z}_{i,t}$}, which is expressed as
\begin{equation}
    \mathbf{f}_{i,t} = \sigma(\mathbf{W}_f \cdot [\mathbf{h}_{i,t-1}, \mathbf{z}_{i,t}] + \mathbf{b}_f).
\end{equation}

Next, a \textit{tanh} function creates a vector of new candidate values {$\mathbf{\tilde{c}}_{i,t}$}, which is expressed as
\begin{equation}
    \mathbf{\tilde{c}}_{i,t} = \tanh(\mathbf{W}_C \cdot [\mathbf{h}_{i,t-1}, \mathbf{z}_{i,t}] + \mathbf{b}_C).
\end{equation}
Then, the current cell state {$\mathbf{c}_{i,t}$} is calculated based on the input gate vector {$\mathbf{s}_{i,t}$}, the forget gate vector {$\mathbf{f}_{i,t}$}, the previous cell state {$\mathbf{c}_{i,t-1}$}, and the value {$\mathbf{\tilde{c}}_{i,t}$}, which is shown as
\begin{equation}\label{ct}
    \mathbf{c}_{i,t} = \mathbf{f}_{i,t} \odot \mathbf{c}_{i,t-1} + \mathbf{s}_{i,t} \odot \mathbf{\tilde{c}}_{i,t}.
\end{equation}
After passing through the output gate, {the output $\mathbf{\hat{z}}_{i,t}$, as well as $\mathbf{h}_{i,t}$}, are finally obtained as
\begin{equation}
    \mathbf{\hat{z}}_{i,t} = \mathbf{h}_{i,t} = \mathbf{o}_{i,t} \odot \tanh(\mathbf{c}_{i,t}).
    \label{eqn:sim_out}
\end{equation}

In our study, we only utilize a single cell unit to form the LSTM structure, which indicates that we do not utilize the recurrent structure originally from LSTM structure.
The latent representation {$\mathbf{z}_{i,t}$} in our study is regarded as time-series data with only one time step.
Therefore, both {$\mathbf{h}_{i,t-1}$ and $\mathbf{c}_{i,t-1}$} are set to the fixed values {$\mathbf{h}_{i,0}$ and $\mathbf{c}_{i,0}$} respectively.
In our study, both {$\mathbf{h}_{i,0}$ and $\mathbf{c}_{i,0}$} are set to zero matrices. In the case of  {$\mathbf{c}_{i,0}=\mathbf{O}$}, \eqref{ct} becomes
\begin{equation}
    \begin{aligned}
         & \mathbf{c}_{i,t} = \mathbf{s}_t \odot \mathbf{\tilde{c}}_{i,t}. \\
    \end{aligned}
    \label{eqn:sim_in}
\end{equation}

The main role of degenerated LSTM module can be described as follows.
Firstly, by treating the training data as a temporal sequence, the data {$\mathbf{\tilde{c}}_{i,t}$} inherently retains information from the initial stage or receive {$\mathbf{z}_{i,t}$} from the encoder in CAE.
Therefore, since the data in the training set are assumed to be normal, the cell unit preserves the prototypical normal patterns in the normal set and stores them into {$\mathbf{\tilde{c}}_{i,t}$}.
{
Secondly, both input and output gates result from the logistic sigmoid function, which limits their values ranging from 0 to 1. In addition, it has also been discovered that the average values of input and output gates in the case of normal inputs are smaller than those in the case of abnormal inputs. Consequently, normal inputs yield small output values according to (12) and (11), resulting in smaller anomaly scores as compared to abnormal inputs. Therefore, the input gate can remove abnormal information while retaining the information of normal samples. Furthermore, the output gate can filter out abnormal information to distinguish anomalous data from normal data in more effective way. In the subsequent experiments of this paper, it can be observed that replacing input and output gates with identity matrices noticeably impairs the functionality of anomaly detection, which substantiates the indispensable importance of input and output gates in enhancing the performance of identifying anomalies. In addition, we have also substituted $\mathbf{\tilde{c}}_{i,t}$ with an identity matrix. Experimental results show that this operation results in poorer performance in detecting anomalies, which further indicates the significance of LSTM module.}

\textit{Remark: Through the above analysis, it is evident that the LSTM module is capable of memorizing normal patterns in the training set. Actually, we have also noticed that the memory module in the MemAE model \cite{gong2019memorizing} also has the function of memorization.
    However, due to the possible presence of abnormal data in the training set, the memory module in \cite{gong2019memorizing} records normal patterns in the memory unit containing a set of vectors. The weight assigned to each vector in the memory unit for choosing the pattern that well matches the input data needs to be calculated. However, the memory module in \cite{gong2019memorizing} has the drawback that any interference in the weights would bias the latent representation from the memory module, which subsequently affects the distinguishing ability of autoencoder structure.
    In contrast, in the LSTM module, both input gate and output gate can filter out the input information flow, due to which the interference of abnormal information flow can be mitigated.
    Furthermore, the cell unit in LSTM neither saves multiple state vectors nor uses weighted summing, thereby enhancing the purity of the normal information flow.}

\subsection{Detailed introduction for the \MODEL{} model}
In this section, we {will} describe our proposed \MODEL{} model in detail.
Considering the batch of input data $\{\mathbf{x}_1,\mathbf{x}_2,...,\mathbf{x}_n\} \in \mathcal{X}$, where $n$ denotes the total number of input samples, the objective of the \MODEL{} model is to distinguish abnormal samples from normal samples in more efficient way by means of LSTM to better separate abnormal samples from normal samples, and KL divergence to avoid feature collapse in SVDD training.

Define $\varphi(\cdot ; \theta_m) :\mathbf{z} \to \mathbf{\hat{z}} $ as the process of LSTM module, where $\mathbf{\hat{z}}$ is the output vector of LSTM module.
The data $\mathbf{\hat{z}}$ are then fed into two other modules, which are the SVDD module and the decoder module respectively.
The role the SVDD module is to enforce normal samples getting closer by constructing a hypersphere with the minimal volume enclosing features vectors of all normal samples.
At the same time, the data $\mathbf{\hat{z}}$ in the hidden space are decoded to generate the reconstructed data via the mapping $f_d(\cdot ; \theta_d) :\mathbf{\hat{z}} \to \mathbf{\hat{x}}$, where $\mathbf{\hat{x}}$ is the output of decoder as the reconstructed data of $\mathbf{x}$.
In addition, we have to mention that in the training process, the data $\mathbf{\hat{z}}$ are enforced following the standard Gaussian distribution by using the penalty of KL divergence.
The use of KL divergence can be helpful for broadening the distribution of output data from LSTM module to mitigate the chance of feature collapse in the SVDD module.

In our work, we optimize the \MODEL{} model by minimizing the total loss $L_{total}$,  which is the weighted sum of Deep SVDD loss $L_{svdd}$, KL divergence $L_{kl}$ between $\mathbf{\hat{z}}$ and standard Gaussian distribution, and the reconstruction loss $L_{rec}$ from the CAE model.
The total loss can be expressed as
\begin{equation}
    L_{total}=L_{svdd}+\lambda_1 L_{kl}+\lambda_2 L_{rec},
    \label{eqn:totalloss}
\end{equation}
where the hyperparameters $\lambda_1$ and $\lambda_2$ are used to trade off different kinds of losses.
For the SVDD loss $L_{svdd}$, we can use either the soft-boundary version $L_{soft}$ or the hard-boundary version $L_{hard}$.

In addition, the reconstruction loss from the CAE module is defined as
\begin{equation}
    \begin{aligned}
        L_{rec} = & \frac{1}{n} \sum_{i=1}^n\|\mathbf{x}_i-\mathbf{\hat{x}}_i\|^2.
    \end{aligned}
    \label{eqn:soft}
\end{equation}

With $L_{svdd}$, $L_{kl}$ and $L_{rec}$, the total loss of the IAE-LSTM-KL model with soft-version of SVDD can be expressed as:
\begin{equation}
    \begin{aligned}
        \min_{\theta_e, \theta_d, R, \mathcal{W}} & R^2+\frac{1}{\upsilon n} \sum_{i=1}^{n} max(0, \|\mathbf{\hat{z}}_i - \mathbf{c}\|^2-R^2)                                                                                                     \\
                                                  & +  \frac{\lambda_1}{n} \sum_{i=1}^{n} K L\left(\mathbf{\hat{z}}_i \|\mathcal{N}_{\mathbf{0},\mathbf{I}} \right)  +   \frac{\lambda_2}{n} \sum_{i=1}^{n} \|\mathbf{x}_i-\mathbf{\hat{x}}_i\|^2 \\
                                                  & +  \frac{\lambda_3}{2}\sum_{l=1}^{L} \|\mathbf{W}^l\|_{F}^2.                                                                                                                                  \\
    \end{aligned}\label{eqn:iae_soft}
\end{equation}
When the soft-boundary of SVDD is replaced by hard-boundary SVDD, the total loss becomes
\begin{equation}
    \begin{aligned}
        \min_{\theta_e, \theta_d, \mathcal{W}} & \frac{1}{n} \sum_{i=1}^{n}\|\mathbf{\hat{z}}_i - \mathbf{c}\|^2 + \frac{\lambda_1}{n}\sum_{i=1}^{n}K L\left(\mathbf{\hat{z}}_i \|\mathcal{N}_{\mathbf{0},\mathbf{I}} \right) \\
                                               & + \frac{\lambda_2}{n}\sum_{i=1}^{n} \|\mathbf{x}_i-\mathbf{\hat{x}}_i\|^2  +  \frac{\lambda_3}{2}\sum_{l=1}^{L} \|\mathbf{W}^l\|_{F}^2.                                      \\
    \end{aligned}\label{eqn:iae_hard}
\end{equation}
In \eqref{eqn:iae_soft} and \eqref{eqn:iae_hard}, the hyperparameter $\lambda_3$ is set to a positive value controlling the degree of decay regularizer.

In the testing stage, we resolve to detect anomalies by measuring the distance between the feature vector of testing sample to the hypersphere center in the hidden feature space, which is defined as
\begin{equation}
    \begin{aligned}
        SCORE = \|\mathbf{\hat{z}}- \mathbf{c}\|^2.
    \end{aligned}
    \label{eqn:score}
\end{equation}
\section{Experiments}
\label{sec:results}
In this section, we {will} evaluate the performance of our proposed \MODEL{} model by comparing it with several other state-of-art models in the experiments for anomaly detection.
Our experiments are carried out on three image datasets and one time-series dataset.
Below we {will} firstly describe the datasets that are used in our experiments.
Then, we {will} present and analyze experimental results of the \MODEL{} model and several other competing models.

\subsection{Datasets}
In our study, we have carried out extensive experiments on three commonly-used image datasets, which are CIFAR10 \cite{krizhevsky2009learning},  Fashion MNIST \cite{xiao2017fashion} and MVTec AD \cite{bergmann2019mvtec} datasets, and one time-series dataset, i.e. the Wind Turbine Blade Icing (WTBI) dataset. Below are the basic introductions for these datasets.
%Both CIFAR10 and Fashion MNIST datasets have multiple classes and are originally used for classification.
%The MVTec AD dataset is a comprehensive collection of defect-free and defective images of various object types and materials, designed to facilitate the development and evaluation of anomaly-detection algorithms in the scenario of industrial inspection.
%In addition, the WTBI dataset is collected from the measurements of wind turbines.
%To apply our proposed and other state-of-art competing methods to the WTBI dataset, we transform the time-series data into the image-like data to detect abnormal events that are hidden in the time-series data. Below are the brief introductions for the above datasets.

%\subsubsection{CIFAR10}
1). \emph{CIFAR10}: The CIFAR10 dataset consists of 60,000 $32 \times 32$ RGB color images categorized into 10 different classes\cite{krizhevsky2009learning}.
The training and testing set consist of 50,000 and 10,000 samples respectively. In each experiment on this dataset, we designate one class as normal and the others as abnormal.

%\subsubsection{Fashion MNIST}
2). \emph{Fashion MNIST}: The Fashion MNIST dataset was created in 2017 and consists of 70,000 grayscale images with $28 \times 28$ pixels forming 10 classes \cite{xiao2017fashion}.
This dataset is divided into two parts with 60,000 images in the training set and 10,000 images in the testing set. As on the CIFAR10 dataset, we also designate one class as normal while considering others as abnormal in each experiment on the Fashion MNIST.

%\SUBSUBSECTION{wIND tURBINE bLADE iCING(wtbi)}
% \footnote{https://github.com/XinArkh/Industrial\_BigData\_Competition\_2017}
3). \emph{WTBI}: The WTBI dataset was collected from the SCADA (Supervisory Control and Data Acquisition) system of wind turbine units.
This dataset consists of SCADA information from multiple turbines within a wind farm, collected over a duration of two months.
In this dataset, there are totally over a million timestamps of SCADA monitoring data. {At each timestamp, a 28-dimension vector is recorded. In each {vector}, the first element is the timestamp index and the last element is the normality/anomaly flag. The remaining 26 elements are monitoring measurements, covering various dimensions such as operational parameters, environmental conditions and status parameters of turbines. Since only the 26 monitoring measurements are useful for detecting anomalies, we ignore the first and the last elements of each 28-dimension vector, forming a 26-dimention vector at each timestamp.} The data in WTBI dataset are divided into two categories, i.e. non-icing period and icing period, with the non-icing and icing periods identified as normal and abnormal states respectively. Since the CAE structure is involved in both our proposed \MODEL{} and other competing state-of-art models, the WTBI dataset is firstly converted into image-like format for further training in anomaly-detection models before training. {To facilitate us applying anomaly-detection models on this dataset, we need to smooth the data between adjacent timestamps. As shown in Fig. \ref{fig:window_smooth}, the original data was firstly preprocessed by taking the average over the previous five timestamps, the current timestamp and the next five timestamps. The window is shifted forward with stride equal to one timestamp for ongoing smoothing processing.} {After smoothing operation, we concatenate 26 consecutive 26-dimensional vectors into a $26 \times 26$ matrix which is subsequently converted into a tensor vector. Then, the 26-timestamp window is moved forward with stride equal to one timestamp for ongoing processing.}
Using the sliding step of one unit, we finally obtained 39,438 normal and 2,814 abnormal image-like samples. The testing set was formed by combing 30\% of normal and all abnormal image-like samples. Finally, we have 27,606 training samples and 14,646 testing samples.
\begin{figure}[!t]
    \centering
    \includegraphics[width=0.85\linewidth]{window_smooth.eps}
    \caption{Illustration of data smoothing based on sliding window.} \label{fig:window_smooth}
    % \includesvg[width=0.9\linewidth]{figures/train_loss_smoothing}
\end{figure}
%\subsubsection{MVTec AD}

4). \emph{MVTec AD}: This dataset contains a carefully curated collection of images spanning a wide range of materials and products \cite{bergmann2019mvtec}, which contain more than 5,000 high-resolution images meticulously categorized into fifteen distinct classes.
Each class includes a specified kind of objects and textures. Within each category, there are pristine training images, a selection demonstrating various defects, and additional defect-free images.
Each image in this dataset is sized with $300 \times 300$ pixels, making it a comprehensive resource for conducting research in anomaly detection.

\subsection{Results}
The Area Under the Receiver Operating Characteristic (AUROC) evaluation has been widely adopted as the benchmark for assessing classification performance in terms of discrimination accuracy \cite{liu2018future,liu2021hybrid, wyatt2022anoddpm}.
This evaluation facilitates straightforward comparisons across different models, as it does not necessitate the selection of classification thresholds or any additional parameters during calculation.
In our study, we have conducted comprehensive comparisons between \MODEL{} and several other state-of-the-art models including CAE \cite{masci2011stacked}, MemAE \cite{gong2019memorizing}, VAE \cite{kingma2013auto}, Gaussian AD \cite{rippel2021gaussian}, LSA \cite{abati2019latentspace}, Deep SVDD \cite{ruff2018deep} and IAEAD \cite{cheng2021improved}.
The original experiments designed for the Gaussian AD method in \cite{rippel2021gaussian} were carried out in supervised manner, which do not meet our expectations. Therefore, we modified the Gaussian AD method in \cite{rippel2021gaussian} to the unsupervised version in our experiments.
In {\cite{rippel2021gaussian}}, the performance of Gaussian AD on the MVTec AD dataset was validated using five-fold cross-validation.
However, in our work, we do not follow the way of dataset split as that in \cite{rippel2021gaussian} since there are no anomalies in the training set. We keep the original split of MVTec AD dataset.

\begin{figure*}[!t]
    \centering
    \includegraphics[width=0.8\linewidth]{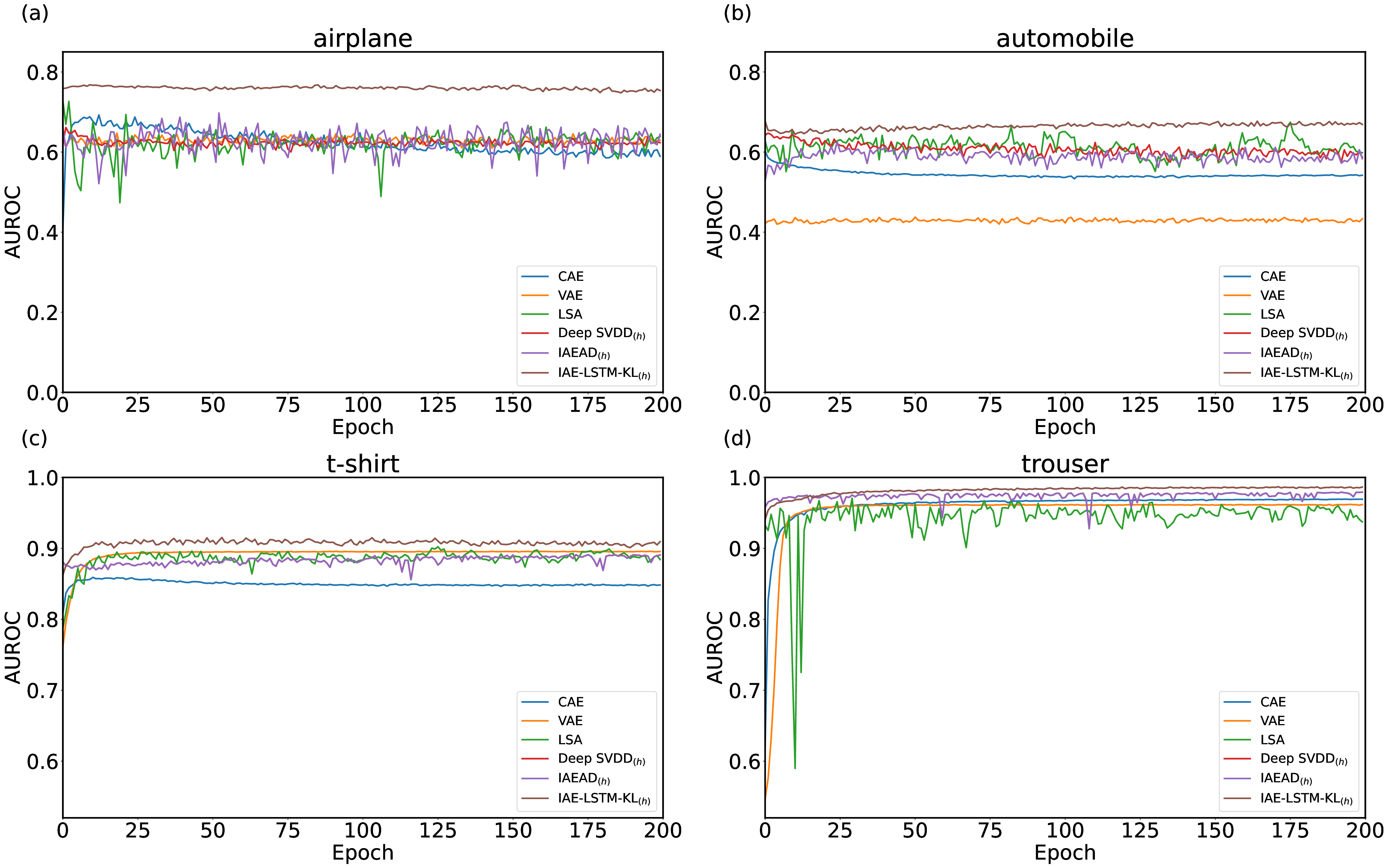}
    \caption{AUROC values \textit{versus} epochs in the training stage on CIFAR10 ((a) and (b)) and Fashion MNIST ((c) and (d)) datasets. The normal class in the dataset is shown above each subfigure. The training set only contains normal samples.} \label{fig:auroc}
\end{figure*}
% \begin{figure}[!t]
%     \centering
%     \includegraphics[width=0.8\linewidth]{wtbi_roc.eps}
%     \caption{AUROC values \textit{versus} epochs in the training stage on the WTBI dataset.} \label{fig:wtbi_roc}
% \end{figure}
\renewcommand{\b}{\bfseries}
\renewcommand{\u}[1]{\underline{#1}}
\newcommand{\m}[1]{{\scriptsize #1}}
\begin{table*}[!t]
    \caption{AUROC on the \cifar{} dataset.}
    \begin{tabularx}{\linewidth}{l>{\centering}X>{\centering}X>{\centering}X>{\centering}X>{\centering}X>{\centering}X>{\centering}X>{\centering}X>{\centering}X>{\centering}XX}
        \toprule
        Class             & airplane  & automobile & bird      & cat       & deer      & dog       & frog      & horse     & ship      & truck     & Avg.      \\
        \midrule
        CAE               & 0.581     & 0.531      & 0.488     & 0.571     & 0.497     & 0.601     & 0.441     & 0.557     & 0.739     & 0.612     & 0.562     \\
        MemAE             & 0.658     & 0.537      & 0.483     & 0.593     & 0.497     & 0.621     & 0.442     & 0.570     & 0.789     & 0.624     & 0.581     \\
        VAE               & 0.629     & 0.433      & \b{0.618} & 0.499     & \b{0.730} & 0.516     & 0.679     & 0.522     & 0.668     & 0.539     & 0.583     \\
        Gaussian AD       & 0.439     & 0.498      & 0.512     & 0.552     & 0.504     & 0.547     & 0.506     & 0.546     & 0.570     & 0.527     & 0.520     \\
        % Pixel CNN\cite{van2016conditional}$^\star$     & 0.403     & 0.428 & 0.574 & 0.547 & 0.580 & 0.648 & 0.746 & 0.659 & 0.705 & 0.660 & 0.551 \\

        % AnoGAN\cite{schlegl2017unsupervised}$^\dagger$ & 0.528     & 0.574 & 0.584 & 0.545 & 0.542 & 0.560 & 0.578 & 0.591 & 0.601 & 0.660 & 0.618 \\

        LSA               & 0.621     & 0.602      & 0.480     & 0.561     & 0.606     & 0.609     & 0.658     & 0.632     & 0.780     & 0.723     & 0.627     \\

        Deep SVDD$_{(h)}$ & 0.621     & 0.620      & 0.486     & 0.576     & 0.569     & 0.633     & 0.578     & 0.614     & 0.777     & 0.688     & 0.616     \\
        Deep SVDD$_{(s)}$ & 0.616     & 0.620      & 0.481     & 0.563     & 0.563     & 0.614     & 0.588     & 0.617     & 0.772     & 0.681     & 0.611     \\

        IAEAD$_{(h)} $    & 0.637     & 0.590      & 0.500     & 0.557     & 0.597     & 0.581     & 0.637     & 0.631     & 0.725     & 0.663     & 0.612     \\
        IAEAD$_{(s)} $    & 0.632     & 0.583      & 0.506     & 0.590     & 0.543     & 0.613     & 0.582     & 0.632     & 0.768     & 0.670     & 0.612     \\

        \hline
        \MODEL$_{(h)}$    & \b{0.780} & \b{0.688}  & 0.613     & 0.648     & 0.706     & 0.650     & \b{0.759} & \b{0.658} & \b{0.798} & \b{0.777} & \b{0.708} \\
        \MODEL$_{(s)}$    & 0.761     & 0.642      & 0.611     & \b{0.670} & 0.703     & \b{0.695} & 0.725     & 0.643     & 0.752     & 0.735     & 0.694     \\

        % \cmidrule{1-9}
        %\cmidrule(r){1-7}\cmidrule(l){8-9} %\cmidrule(lr){19-19}

        %Mean & 64.8  $\pm$ .02.5 & 64.9  $\pm$ .02.5 & 59.5  $\pm$ .02.4 & 61.8  $\pm$ .02.1 & $63.4 \pm 2.2$ & \b \u{67.6}  $\pm$ .02.4 & 64.8  $\pm$ .02.2 & \b 66.9  $\pm$ .02.3 \\

        \bottomrule

        \multicolumn{9}{@{}l}{}
    \end{tabularx}
    \label{tab:cifar10}
\end{table*}

\begin{table*}[!t]
    \caption{AUROC on the Fashion MNIST dataset.}
    \begin{tabularx}{\linewidth}{l>{\centering}X>{\centering}X>{\centering}X>{\centering}X>{\centering}X>{\centering}X>{\centering}X>{\centering}X>{\centering}XcX}
        \toprule
        Class             & t-shirt   & trouser   & pullover  & dress     & coat      & sandal    & shirt     & sneaker   & bag       & ankle-boot & Avg.      \\
        \midrule
        CAE               & 0.846     & 0.966     & 0.813     & 0.855     & 0.890     & 0.571     & 0.728     & 0.942     & 0.838     & 0.984      & 0.843     \\
        MemAE             & 0.901     & 0.987     & 0.885     & 0.916     & 0.883     & 0.898     & 0.782     & 0.988     & 0.861     & 0.980      & 0.908     \\
        VAE               & 0.896     & 0.962     & 0.863     & 0.884     & 0.894     & 0.347     & 0.773     & 0.968     & 0.853     & 0.983      & 0.842     \\
        Gaussian AD       & 0.596     & 0.834     & 0.607     & 0.671     & 0.541     & 0.366     & 0.507     & 0.577     & 0.629     & 0.524      & 0.585     \\
        LSA               & 0.886     & 0.940     & 0.862     & 0.867     & 0.895     & 0.474     & 0.766     & 0.941     & 0.916     & 0.984      & 0.853     \\

        Deep SVDD$_{(h)}$ & 0.860     & 0.980     & 0.836     & 0.872     & 0.892     & 0.684     & 0.758     & 0.948     & 0.934     & 0.973      & 0.874     \\
        Deep SVDD$_{(s)}$ & 0.863     & 0.977     & 0.851     & 0.882     & 0.891     & 0.649     & 0.775     & 0.951     & 0.923     & 0.975      & 0.874     \\

        IAEAD$_{(h)}$     & 0.878     & 0.941     & 0.871     & 0.853     & 0.907     & 0.543     & 0.781     & 0.955     & 0.891     & 0.986      & 0.861     \\
        IAEAD$_{(s)}$     & 0.864     & 0.961     & 0.851     & 0.856     & 0.895     & 0.692     & 0.759     & 0.953     & 0.879     & 0.986      & 0.870     \\

        \hline
        \MODEL$_{(h)}$    & \b{0.915} & \b{0.987} & \b{0.905} & \b{0.945} & \b{0.928} & 0.874     & 0.814     & \b{0.990} & \b{0.955} & \b{0.992}  & \b{0.931} \\
        \MODEL$_{(s)}$    & 0.887     & 0.981     & 0.891     & 0.915     & 0.912     & \b{0.923} & \b{0.815} & 0.973     & 0.944     & 0.981      & 0.922     \\

        \bottomrule

        \multicolumn{9}{@{}l}{}
    \end{tabularx}
    \label{tab:fmnist}
\end{table*}

\begin{table}[!ht]
    \caption{AUROC on the WTBI dataset.
        The subscripts $(s)$ and $(h)$ refer to the soft-boundary and hard-boundary SVDD, respectively.
        We emphasize in bold the performance of the best models.}
    \begin{tabularx}{\linewidth}{>{\centering}X>{\centering\arraybackslash}X}

        Models            & AUROC     \\
        \midrule

        CAE               & 0.704     \\
        MemAE             & 0.690     \\
        VAE               & 0.392     \\
        Gaussian AD       & 0.595     \\
        LSA               & 0.685     \\
        Deep SVDD$_{(h)}$ & 0.798     \\
        Deep SVDD$_{(s)}$ & 0.823     \\
        IAEAD$_{(h)}$     & 0.823     \\
        IAEAD$_{(s)}$     & 0.894     \\
        \hline
        \MODEL$_{(h)}$    & 0.893     \\
        \MODEL$_{(s)}$    & \b{0.963} \\
        \bottomrule

        \multicolumn{2}{@{}l}{}       \\
    \end{tabularx}

    \label{tab:wtbi}

\end{table}
\begin{table*}[!t]
    \caption{AUROC on the MVTec AD dataset.}
    \begin{tabularx}{\linewidth}{lccccccccccc}
        \toprule
        Class          & CAE       & MemAE     & VAE   & Gaussian AD & LSA   & Deep SVDD$_{(h)}$ & Deep SVDD$_{(s)}$ & IAEAD$_{(h)}$ & IAEAD$_{(s)}$ & Ours$_{(h)}$ & Ours$_{(s)}$ \\
        \midrule
        \m{Bottle}     & \b{0.980} & 0.973     & 0.506 & 0.678       & 0.888 & 0.901             & 0.925             & 0.883         & 0.886         & 0.940        & 0.945        \\
        \m{Cable}      & 0.768     & 0.753     & 0.595 & 0.517       & 0.724 & 0.839             & 0.843             & 0.673         & 0.673         & \b{0.856}    & 0.850        \\
        \m{Capsule}    & 0.619     & 0.627     & 0.523 & 0.563       & 0.555 & \b{0.753}         & 0.616             & 0.512         & 0.529         & 0.713        & 0.716        \\
        \m{Carpet}     & 0.532     & 0.546     & 0.409 & 0.637       & 0.486 & 0.704             & 0.679             & 0.569         & 0.581         & \b{0.800}    & 0.666        \\
        \m{Grid}       & 0.810     & 0.800     & 0.643 & 0.437       & 0.749 & 0.887             & 0.505             & 0.776         & 0.804         & \b{0.906}    & 0.904        \\
        \m{Hazelnut}   & 0.860     & \b{0.860} & 0.595 & 0.353       & 0.602 & 0.718             & 0.734             & 0.667         & 0.623         & 0.756        & 0.757        \\
        \m{Leather}    & 0.477     & 0.748     & 0.416 & \b{0.976}   & 0.402 & 0.615             & 0.492             & 0.792         & 0.647         & 0.571        & 0.662        \\
        \m{Metal nut}  & 0.569     & 0.541     & 0.272 & 0.503       & 0.570 & \b{0.877}         & 0.853             & 0.323         & 0.489         & 0.735        & 0.761        \\
        \m{Pill}       & 0.759     & 0.759     & 0.441 & 0.525       & 0.703 & \b{0.808}         & 0.806             & 0.643         & 0.643         & 0.768        & 0.758        \\
        \m{Screw}      & 0.482     & 0.707     & 0.522 & 0.480       & 0.038 & 0.137             & 0.093             & 0.328         & 0.285         & \b{0.963}    & 0.869        \\
        \m{Tile}       & 0.582     & 0.541     & 0.708 & 0.723       & 0.678 & 0.740             & 0.738             & 0.758         & 0.701         & 0.723        & \b{0.796}    \\
        \m{Toothbrush} & 0.985     & 0.988     & 0.357 & 0.381       & 0.923 & 1.000             & 0.997             & 0.902         & 0.899         & \b{1.000}    & 1.000        \\
        \m{Transistor} & 0.801     & 0.771     & 0.603 & 0.438       & 0.741 & 0.876             & 0.847             & 0.720         & 0.763         & 0.889        & \b{0.900}    \\
        \m{Wood}       & 0.914     & 0.910     & 0.922 & 0.713       & 0.924 & 0.757             & 0.850             & \b{0.969}     & 0.938         & 0.927        & 0.959        \\
        \m{Zipper}     & 0.871     & 0.853     & 0.360 & 0.651       & 0.657 & 0.872             & 0.798             & 0.792         & 0.564         & 0.825        & \b{0.877}    \\
        \m{Avg.}       & 0.734     & 0.759     & 0.525 & 0.572       & 0.643 & 0.766             & 0.718             & 0.687         & 0.668         & 0.825        & \b{0.828}    \\

        \bottomrule

        % \multicolumn{9}{@{}l}{}
    \end{tabularx}
    \label{tab:mvtec}
\end{table*}

% We utilize the notation Conv2($s_{in}$, $s_{out}$, $k$) and DConv2($s_{in}$, $s_{out}$, $k$) to denote the two-dimensional convolutional and deconvolutional layers respectively, with $s_{in}$, $s_{out}$, and $k$ representing input channel size, output channel size, and kernel size respectively.
We utilize the notation Conv2($s_{in}$, $s_{out}$, $k$) and {DConv2($s_{in}$, $s_{out}$, $k$)} to denote the two-dimensional convolutional and deconvolutional layers respectively, with $s_{in}$, $s_{out}$, and $k$ representing input channel size, output channel size, and kernel size respectively.
In addition, a linear layer is denoted by Linear($n_{in}$, $n_{out}$), where $n_{in}$ and $n_{out}$ represent the number of input and output features respectively.
Similarly, we use the notation LSTM($n_{in}$, $n_{out}$) to represent the LSTM layer, with $n_{in}$ and $n_{out}$ also represent the number of input and output features respectively.

For CIFAR10, Fashion MNIST and WTBI datasets, we use three convolutional layers to implement the encoder: Conv2($c$, 32, 5)-Conv2(32, 64, 5)-Conv2(64, 128, 5), where $c$ indicates the number of image channels.
The decoder is implemented as Dconv2(8,128,5)-Dconv2(128, 64, 5)-Dconv2(64, 32, 5)-Dconv2(32, $c$, 5).
Since only the CIFAR10 dataset contains the colored images, we set $c=3$ for the CIFAR10 dataset.
For the remaining two datasets, i.e. Fashion MNIST and WTBI datasets, we set $c=1$.
Besides, each layer of the autoencoder is followed by batch normalization \cite{ioffe2015batch} and a leaky ReLU activation \cite{maas2013rectifier}, except for the last Dconv2 layer which is appended by a sigmoid function.
Additionally, a max pooling layer is included following each activation function in the encoder, with both kernel size and stride set to 2.
For the LSTM module, we use two layers, i.e. LSTM(2048, 2048)-Linear(2048, 128).
    {Then, the output from LSTM module is shaped into $8 \times 4 \times 4$ as the input into the decoder.}
Referring to the MVTec AD dataset, we use encoder and decoder with higher capacities. For the encoder, we use Conv2(3,96,3)-Conv2(96,128,3)-Conv2(128,256,3)-Conv2(256,256,3), while for the decoder, we use Dconv2(256,256,3)-Dconv2(256,128,3)-Dconv2(128,96,3)-Dconv2(96,3,3).
Each layer is also followed by batch normalization and a leaky ReLU activation, except for the last Dconv2 layer.
Finally, the LSTM module is implemented as LSTM(4096, 4096),
{from which the output is then reshaped into $256 \times 4 \times 4$ as the input into the decoder.}

    {All experiments in our study are implemented based on Adam optimizer \cite{kingma2014adam} and PyTorch Lightning framework \cite{Falcon_PyTorch_Lightning_2019}. The weights in all models under study are optimized using stochastic gradient algorithm implemented by the Adam optimizer. The hyperparameters, i.e.  $\lambda_1$ and $\lambda_2$ in \eqref{eqn:totalloss}, are adjusted separately for each class within the dataset to be normal via the Optuna framework  \cite{optuna_2019} using the grid search strategy. Then, for each model under study, the optimal values of hyperparameters are selected to compare with other models with their respective optimal hyperparameter values.}

%{This experiment is implemented using Adam optimizer \cite{kingma2014adam} and PyTorch Lightning framework \cite{Falcon_PyTorch_Lightning_2019}.
%    The weights of model units are optimized by the stochastic gradient algorithm which has been implemented by the Adam optimizer.
%    $\lambda_1$ and $\lambda_2$ are adjusted separately for each class within the dataset to be normal via Optuna framework \cite{optuna_2019} with the grid search strategy.}

Comparing results for all competing models on CIFAR10, Fashion MNIST, WTBI and MVTec AD datasets  are demonstrated in Table~\ref{tab:cifar10}, Table~\ref{tab:fmnist}, Table~\ref{tab:wtbi} and Table~\ref{tab:mvtec} respectively.
The subscripts $(s)$ and $(h)$ refer to the soft-boundary and hard-boundary SVDD, respectively. AUROC values of top-performing models are shown in bold.
It can be observed from these results that the proposed \MODEL{} model exhibits higher average AUROC values than other state-of-art models.
Moreover, we have also observed that in the \MODEL{} model, the performances of anomaly detection are basically in the same level when using soft-boundary and hard-boundary SVDD module respectively.
These results show that our proposed \MODEL{} model is superior to the competing models in anomaly detection.

\begin{figure}[!t]
    \centering
    \includegraphics[width=0.7\linewidth]{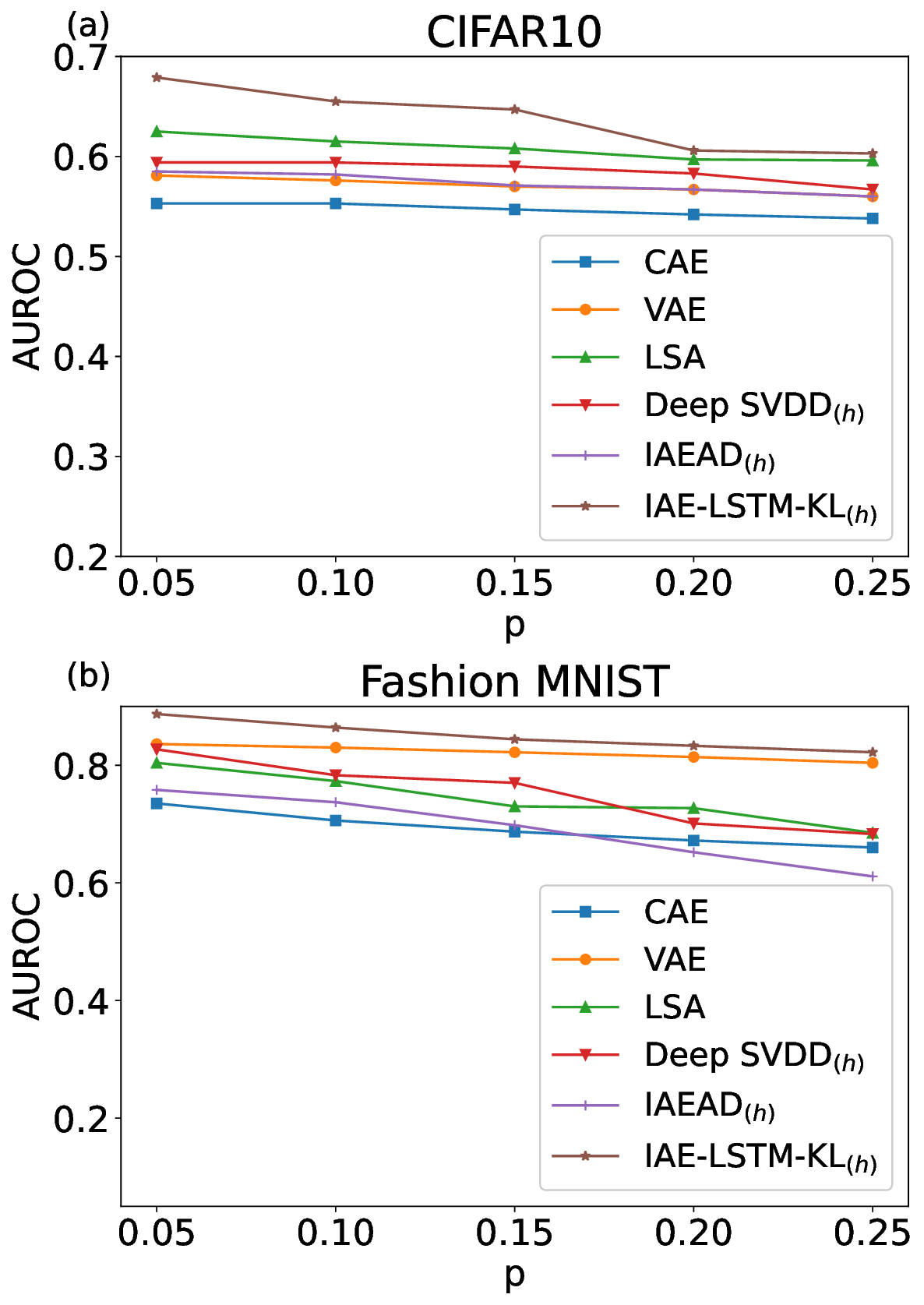}
    \caption{Relationship between AUROC values and anomaly ratio on the training set. (a). CIFAR10 dataset. (b). Fashion MNIST dataset.} \label{fig:radio}
\end{figure}

Fig.~\ref{fig:auroc} demonstrates the change of AUROC values of all models in the training stage on CIFAR10 and Fashion MNIST datasets.
Firstly, it can be observed that AUROC values of the \MODEL{} model consistently reach high levels from the outset of training process thanks to the pretraining operation before the formal training for the \MODEL{} model.
Then, we have also found that AUROC values of the \MODEL{} model stabilize at the highest level among all models under study from the initial stage of training process.
Moreover, AUROC values of the \MODEL{} model fluctuate only slightly in the steady-state on all datasets under study.
However, the stability performances of some other models are not so pronounced as that of the \MODEL{} model.
For example, dramatic fluctuations of AUROC values in LSA model are observed on both datasets.
    {Finally, we have also noticed that AUROC values of IAE-LSTM-KL model is capable of going towards stability in very few iterations when training on both CIFAR10 and Fashion MNIST datasets. On the contrary, some other methods, for example the VAE model on Fashion MNIST datasets, need more iterations to reach the steady-state AUROC values. This result well corroborates the faster convergence of our proposed IAE-LSTM-KL model in training.}

In order to evaluate resilience of the \MODEL{} model in the presence of increased abnormal samples within the dataset, we manipulate the proportion of abnormal samples and replicate the aforementioned experiments on CIFAR10 and Fashion MNIST datasets.
In each dataset, the abnormal data samples are randomly chosen from the remaining nine categories. Then, the chosen abnormal data samples are added into the normal samples to form the augmented training sets.
Abnormal ratios $\rho$ in the augmented training set are set to 5\%, 10\%, 15\%, 20\%, and 25\%, respectively.
The impact of abnormal ratios on AUROC for all competing models in our study is presented in Fig.~\ref{fig:radio}.
To illustrate the statistical findings across all models, we calculate the average AUROC values over all ten classes on each dataset.
Fig.~\ref{fig:radio} reveals some interesting results.
Firstly, with an increase of abnormal samples in the training set, downward trends in the average AUROC values are observed for all models under study.
This observation is straightforward, as a higher anomaly ratio in the training set introduces more noises into the training process, ultimately leading to degradation in the performance of anomaly detection.
More importantly, our proposed \MODEL{} model consistently demonstrates higher AUROC values than other competing models, regardless of the anomaly level in the training set.
This result suggests that the \MODEL{} model exhibits superior anomaly-detection capabilities across varying levels of noise in the training set when compared to other competing models.

\subsection{Ablation Studies}
\begin{table*}[!t]
    \caption{Ablation studies for the IAE-LSTM-KL model on the the \cifar{} dataset. }
    \begin{tabularx}{\linewidth}{l>{\centering}X>{\centering}X>{\centering}X>{\centering}X>{\centering}X>{\centering}XXXXXX}
        \toprule
        Class            & airplane  & automobile & bird      & cat       & deer      & dog       & frog      & horse     & ship      & truck     & Avg.      \\
        \midrule
        \ABLSDG$_{(h)}$  & 0.627     & 0.626      & 0.546     & 0.544     & 0.616     & 0.608     & 0.542     & 0.624     & 0.732     & 0.671     & 0.614     \\
        \ABLSDG$_{(s)}$  & 0.629     & 0.638      & 0.562     & 0.542     & 0.595     & 0.601     & 0.589     & 0.618     & 0.741     & 0.667     & 0.618     \\
        \ABLSDL$_{(h)}$  & 0.603     & 0.634      & 0.486     & 0.610     & 0.579     & 0.644     & 0.618     & 0.647     & 0.775     & 0.723     & 0.632     \\
        \ABLSDL$_{(s)}$  & 0.605     & 0.638      & 0.495     & 0.610     & 0.573     & 0.639     & 0.621     & 0.649     & 0.778     & 0.722     & 0.633     \\
        \ABLSDLG$_{(h)}$ & 0.689     & 0.642      & 0.528     & 0.604     & 0.585     & 0.638     & 0.625     & 0.643     & 0.782     & 0.722     & 0.646     \\
        \ABLSDLG$_{(s)}$ & 0.620     & 0.642      & 0.521     & 0.611     & 0.603     & 0.638     & 0.632     & 0.650     & 0.780     & 0.712     & 0.641     \\
        %\hline
        % DA*SVDD
        \ABLAEG$_{(h)}$  & 0.764     & 0.659      & 0.607     & 0.649     & \b{0.711} & 0.680     & 0.729     & 0.658     & \b{0.817} & 0.766     & 0.704     \\
        \ABLAEG$_{(s)}$  & 0.768     & 0.581      & 0.538     & 0.600     & 0.629     & 0.629     & 0.679     & 0.605     & 0.777     & 0.704     & 0.651     \\
        \ABLAEL$_{(h)}$  & 0.667     & 0.680      & 0.504     & 0.633     & 0.626     & 0.636     & 0.719     & 0.645     & 0.793     & 0.760     & 0.666     \\
        \ABLAEL$_{(s)}$  & 0.708     & 0.644      & 0.549     & 0.593     & 0.625     & 0.656     & 0.663     & 0.622     & 0.767     & 0.712     & 0.654     \\
        \MODEL$_{(h)}$   & \b{0.780} & \b{0.688}  & \b{0.613} & 0.648     & 0.706     & 0.650     & \b{0.759} & \b{0.658} & 0.798     & \b{0.777} & \b{0.708} \\
        \MODEL$_{(s)}$   & 0.761     & 0.642      & 0.611     & \b{0.670} & 0.703     & \b{0.695} & 0.725     & 0.643     & 0.752     & 0.735     & 0.694     \\

        % \cmidrule{1-9}
        %\cmidrule(r){1-7}\cmidrule(l){8-9} %\cmidrule(lr){19-19}

        %Mean & 64.8  $\pm$ .02.5 & 64.9  $\pm$ .02.5 & 59.5  $\pm$ .02.4 & 61.8  $\pm$ .02.1 & $63.4 \pm 2.2$ & \b \u{67.6}  $\pm$ .02.4 & 64.8  $\pm$ .02.2 & \b 66.9  $\pm$ .02.3 \\

        \bottomrule

        \multicolumn{9}{@{}l}{}
    \end{tabularx}
    \label{tab:cifar10_abl}
\end{table*}

\begin{table*}[!t]
    \caption{Ablation studies for the IAE-LSTM-KL model on the Fashion MNIST dataset.}
    \begin{tabularx}{\linewidth}{l>{\centering}X>{\centering}X>{\centering}X>{\centering}X>{\centering}X>{\centering}X>{\centering}X>{\centering}X>{\centering}XcX}
        \toprule
        Class            & t-shirt   & trouser   & pullover  & dress     & coat      & sandal    & shirt     & sneaker   & bag       & ankle-boot & Avg.      \\
        \midrule
        % D*SVDD
        \ABLSDG$_{(h)}$  & 0.891     & 0.978     & 0.891     & 0.904     & 0.915     & 0.829     & 0.804     & 0.974     & 0.905     & 0.991      & 0.908     \\
        \ABLSDG$_{(s)}$  & 0.887     & 0.978     & 0.890     & 0.902     & 0.916     & 0.889     & 0.804     & 0.971     & 0.911     & 0.991      & 0.914     \\
        \ABLSDL$_{(h)}$  & 0.887     & 0.981     & 0.892     & 0.909     & 0.899     & 0.869     & 0.805     & 0.972     & 0.942     & 0.978      & 0.913     \\
        \ABLSDL$_{(s)}$  & 0.884     & 0.980     & 0.888     & 0.918     & 0.908     & 0.903     & 0.813     & 0.973     & 0.934     & 0.981      & 0.918     \\
        \ABLSDLG$_{(h)}$ & 0.884     & 0.982     & 0.895     & 0.912     & 0.903     & 0.873     & 0.809     & 0.971     & 0.948     & 0.981      & 0.916     \\
        \ABLSDLG$_{(s)}$ & 0.887     & 0.981     & 0.891     & 0.915     & 0.912     & 0.923     & 0.815     & 0.973     & 0.944     & 0.981      & 0.922     \\
        % \hline
        % DA*SVDD
        \ABLAEG$_{(h)}$  & 0.899     & 0.979     & 0.892     & 0.912     & 0.922     & 0.838     & 0.808     & 0.989     & 0.936     & 0.991      & 0.917     \\
        \ABLAEG$_{(s)}$  & 0.906     & 0.971     & 0.868     & 0.906     & 0.913     & 0.841     & 0.809     & 0.984     & 0.921     & \b{0.994}  & 0.911     \\
        \ABLAEL$_{(h)}$  & 0.908     & 0.987     & 0.898     & 0.942     & \b{0.929} & 0.873     & 0.808     & 0.989     & 0.952     & 0.990      & 0.928     \\
        \ABLAEL$_{(s)}$  & \b{0.920} & 0.972     & 0.865     & 0.924     & 0.910     & 0.909     & 0.782     & 0.980     & 0.915     & 0.987      & 0.916     \\
        \MODEL$_{(h)}$   & 0.915     & \b{0.987} & \b{0.905} & \b{0.945} & 0.928     & 0.874     & 0.814     & \b{0.990} & \b{0.955} & 0.992      & \b{0.931} \\
        \MODEL$_{(s)}$   & 0.887     & 0.981     & 0.891     & 0.915     & 0.912     & \b{0.923} & \b{0.815} & 0.973     & 0.944     & 0.981      & 0.922     \\

        % \cmidrule{1-9}
        %\cmidrule(r){1-7}\cmidrule(l){8-9} %\cmidrule(lr){19-19}

        %Mean & 64.8  $\pm$ .02.5 & 64.9  $\pm$ .02.5 & 59.5  $\pm$ .02.4 & 61.8  $\pm$ .02.1 & $63.4 \pm 2.2$ & \b \u{67.6}  $\pm$ .02.4 & 64.8  $\pm$ .02.2 & \b 66.9  $\pm$ .02.3 \\

        \bottomrule

        \multicolumn{9}{@{}l}{}
    \end{tabularx}
    \label{tab:fmnist_abl}
\end{table*}

\begin{table}[!t]
    \caption{AUROC values of LSTM ablation studies on the Fashion MNIST dataset.}
    \begin{tabularx}{\linewidth}{l>{\centering\arraybackslash}X}

        Models                                    & AUROC     \\
        \midrule

        \MODEL$_{(h)}$ w/o input gate             & 0.926     \\
        \MODEL$_{(h)}$ w/o output gate            & 0.922     \\
        \MODEL$_{(h)}$ w/o input and output gates & 0.866     \\
        \MODEL$_{(h)}$                            & \b{0.927} \\
        \bottomrule

        \multicolumn{2}{@{}l}{}                               \\
    \end{tabularx}

    \label{tab:lstm_fmnist}

\end{table}
%%%%%%%%%%%%%%%%%%%%%%%%%%%%%%%%%%%%%%%%%%%%%%%%% %%%%%%%%%%%%%%%%%%%%%%%%%%%%%%%%%%%%%%%%%%%%%%%%% 
%%%%%%%%%%%%%% MVTec maxBA and AUC %%%%%%%%%%%%%%
%%%%%%%%%%%%%%%%%%%%%%%%%%%%%%%%%%%%%%%%%%%%%%%%% %%%%%%%%%%%%%%%%%%%%%%%%%%%%%%%%%%%%%%%%%%%%%%%%% 

To further investigate the role of each module in the \MODEL{} model on the performance of anomaly detection, we have also implemented ablation studies for the \MODEL{} models.
These experiments aim to discern the individual contribution and impact of each module on the performance of \MODEL{} model, providing insights into the effectiveness of such enhancements in deep learning architecture.
The notation of each model under comparison with the \MODEL{} model is described as following:

\begin{itemize}
    \item DSVDD-KL: Compared with the \MODEL{} model, this model does not include the LSTM module.
          In addition, the decoder part in CAE is also not included in this model.
          Therefore, the objective function of this model only contains SVDD loss and KL-divergence loss.
    \item DLSVDD: Compared with the \MODEL{} model, this model does not include the decoder part in CAE.
          In addition, due to the absence of KL divergence, the latent feature vector after the encoder is not forced to follow Gaussian distribution.
          The objective function of this model only contains the SVDD loss.
    \item DLSVDD-KL: Compared with the \MODEL{} model, this model only does not include the decoder part in CAE. In other words, the DLSVDD-KL model is formed by restricting the latent feature vector after LSTM module in DLSVDD model following the Gaussian distribution. The objective function of this model contains the SVDD loss and KL-divergence loss.
    \item IAE-KL: The only difference of this model from the \MODEL{} model is that this model does {not} include the LSTM module.
    \item IAE-LSTM: The only difference of this model from the \MODEL{} model is that this model does {not} incorporate KL divergence to force the latent feature vector following standard Gaussian distribution.
\end{itemize}

Experiments have been implemented to study the importance of each module by leveraging different combinations and permutations of the CAE module, the SVDD module, the LSTM module and the operation of KL divergence.
Comparing results among the \MODEL{} model and its competing counterparts on CIFAR10 and Fashion MNIST are demonstrated in Table~\ref{tab:cifar10_abl} and Table~\ref{tab:fmnist_abl} respectively. We find that in most cases, the \MODEL{} model yields the highest AUROC values than other models with some module (or modules) removed from the \MODEL{} model.
The results indicate that the combination of SVDD module, CAE module, LSTM module and KL divergence can indeed enhance the performance of anomaly detection. Ablation experiments have also been carried out on WTBI and MVTec AD datasets and the superiority of the IAE-LSTM-KL model over other competing model can still be observed. However, due to the limited space, we do not show the experimental results here.

    {Finally, to validate the important role of both input and output gates in LSTM module, we have also investigated performances of models when input gate or/and output gate are removed away from LSTM module in our proposed IAE-LSTM-KL model. The results are shown in Table VII, from which we can clearly observe that removing both input and out gates can seriously degrade the performance of anomaly detection. This result well highlights the crucial significance of input and output gates in the task of anomaly detection.}

\section{Conclusion}
\label{sec:conclusion}
In this paper, we propose a novel anomaly-detection model, called \MODELEXT{} (\MODEL{}) model, to further enhance the performance of anomaly detection.
This model adds an LSTM module between the encoder and the decoder to memorize feature representations of normal data.
In addition, an SVDD module is connected to the output of LSTM module to jointly train the whole \MODEL{} model.
The Kullback-Leibler (KL) divergence regulating the input data of SVDD module following standard Gaussian distribution is used to alleviate the chance of hypersphere collapse in SVDD module.

Extensive experiments have been carried out to compare the performance of the proposed \MODEL{} model with a few other state-of-art models on both synthetic and real-world datasets.
Results show that the proposed \MODEL{} model yields higher detection precision for anomalies.
Besides, we have also found that even though when the training set is contaminated with outliers, the detection rate for anomalies of the \MODEL{} model can still stabilize in the higher level than other competing models. Finally, some ablations studies have been also carried out to validate the significance of each module in the IAE-LSTM-KL model.
In a word, both the effectiveness and robustness of anomaly-detection can be enhanced by using the \MODEL{} model.

\bibliographystyle{IEEEtran}
\bibliography{refs}

\vspace{11pt}

\vfill
\end{document}